# An Interpretable Framework Applying Protein Words to Predict Protein–Small Molecule Complementary Pairing Rules

Jingke Chen[1*], Jingrui Zhong[1*], Tazneen Hossain Tani[1*], Zidong Su[1*], Xiaochun Zhang[1*], Boxue Tian[1]

[1] *MOE Key Laboratory of Bioinformatics, State Key Laboratory of Molecular Oncology, Beijing Frontier Research Center for Biological Structure, School of Pharmaceutical Sciences, Tsinghua University, Beijing, 100084, China*

\* These authors contribute equally to this work.

[#] To whom correspondence should be addressed:

Boxue Tian: boxuetian@mail.tsinghua.edu.cn

**Abstract**

Despite the high accuracy of 'black box' deep learning models, drug discovery still relies on protein-ligand interaction principles and heuristics. To improve interpretability of protein-small molecule binding predictions, we developed the PWRules framework, which applies binding affinity data to identify privileged small molecule fragments and subsequently defines complementary pairing rules between these fragments and protein words (semantic sequence units) through an interpretability module. The resulting word-fragment rules are then ranked by the PWScore function to prioritize active compounds. Evaluations on benchmark datasets show that PWScore achieves competitive performance comparable to the physics-based model (Glide) and the deep learning model (PSICHIC) and shows broad applicability for protein targets outside the training dataset, e.g., SARS-CoV-2 main protease. Notably, PWScore captures complementary interaction information, yielding superior enrichment performance when integrated with these established methods. Structural analysis of protein-ligand complexes indicates that learned word–fragment rules are significantly enriched near ligand-binding pockets, despite training without explicit structural guidance. By extracting and applying complementary pairing rules, PWRules provides an interpretable framework for drug discovery.

**INTRODUCTION**

Drug discovery relies on the physical principles governing protein-ligand interactions, including hydrogen-bond donor–acceptor pairing, hydrophobic enclosure of nonpolar groups, π–π stacking between aromatic rings, and electrostatic complementarity between protein residues and small molecules[1]. In practice, however, application of these principles relies heavily on accumulated experimental knowledge and experience-driven heuristics developed by medicinal chemists[2-4]. Such heuristic approaches guide the design of high-potential drug candidates by promoting favorable interactions while avoiding steric clashes and electrostatic incompatibilities between target proteins and ligand scaffolds[5]. Alternatively, structure-based drug design can provide a framework for applying heuristic knowledge, including pharmacophore modeling[6] and fragment-based drug design[7,8]. Nevertheless, the accessible chemical space and diversity of therapeutic targets is continually expanding, limiting the inherent scalability of approaches that primarily depend on manual curation and heuristics[9,10].

In recent years, the increasing adoption of machine learning and deep learning methods to support various stages of drug discovery has complemented human expertise by automating and accelerating tasks such as virtual screening, binding-site and binding-affinity prediction, and molecular generation[11-17]. At the same time, increasing availability of large-scale molecular datasets supports the development of data-driven models that capture complex, nonlinear relationships which exceed the capacity of simple heuristic rules. Sequence-based methods have also gained attention due to their capability to infer protein–ligand interactions using only protein sequences and molecular representations in the absence of experimentally resolved structures[18-21]. For example, models such as DeepDTA[22] and TransformerCPI[23] have demonstrated strong predictive performance in estimating protein–ligand binding affinity using only sequence-level inputs.

Decades of medicinal chemistry research have shown that protein–ligand recognition is strongly influenced by local interactions between specific active-site residues and small molecule pharmacophores. Despite their predictive accuracy, deep learning models function as a black box, and thus provide limited insight into the molecular determinants underlying their predictions[24]. Although recent prediction frameworks, such as DeepAffinity, have applied attribution-based analyses to probe learned representations in models[25-27], these approaches characterize whole protein or whole ligand interactions, thus limiting the resolution of principles governing local binding interactions, which could be otherwise leveraged by medicinal chemists. This lack of explainability presents a growing need for learning frameworks that surpass predictive performance to provide interpretable representations relevant to protein–ligand recognition. That is, models capable of associating molecular fragments with compatible local sequence patterns offer a promising avenue for bridging data-driven learning with chemically intuitive reasoning in drug discovery.

In this study, we introduce PWRules, a framework for automatic discovery of interpretable complementary pairing rules between protein sequences (protein "words") and small molecule fragments. Based on binding affinity data, PWRules uses small molecule fragments exhibiting privileged association with a given protein to guide model training. The predictions are subsequently translated into interpretable binding rules through an Integrated Gradients (IG)-based interpretability module. We further developed the PWScore function, built on word-fragment rules, to prioritize active compounds. This scoring system demonstrated performance comparable to other ranking systems in active molecule enrichment on benchmark datasets. Notably, by capturing complementary interaction information, its integration with other established methods yields enhanced enrichment performance. PWScore also exhibits generalizability to targets absent in the training set, such as SARS-CoV-2 main protease (M$^{pro}$). Our analyses of protein-ligand complex structure showed that, despite lacking explicit structural guidance, the learned protein word–fragment rules are enriched near ligand-binding sites. Overall, PWRules thus provides an interpretable framework for advancing drug discovery.

**RESULTS**

**An interpretable protein word-based framework for predicting protein–small-molecule complementary pairing rules**

To discover the rules governing complementary pairing between proteins and small molecules, we developed the PWRules framework for identifying small molecule fragments associated with specific protein words[28], i.e., contiguous or discontinuous segments of 5–20 amino acid residues that represent putative functional units. Using protein–small-molecule binding data from PDBbind[29], BindingDB[30], BindingNet[31], and ChEMBL[32] as inputs, PWRules outputs word-fragment pairing rules consisting of protein words individually paired with small molecule fragments. For a given target protein, potential ligands were first analyzed using a refined fragment library wherein each protein–ligand complex in the training dataset with an experimentally verified binding affinity <10 μM was defined as a binding interaction, and those fragments present in >50% of binding-positive ligands designated as "privileged". Using protein-word embeddings as inputs for a supervised Transformer-based encoder architecture, PWRules provides multidimensional vectors, with each dimension representing the predicted probability that a given small molecule fragment is privileged for the input protein. This formulation is analogous to a multi-label protein function prediction task, which we previously demonstrated as an effective approach for extracting high-quality rules for protein words.[28]

Privileged fragments are labeled as 1, while all other fragments are labeled as 0 or NA, thus serving as the supervisory signals for model training. After training and validation, PWRules can automatically extract complementary pairing rules through an Integrated Gradients (IG)-based interpretability analysis. To generate rules for each protein, the model selects only positive (i.e., privileged, label = 1) small molecule fragments with predicted labels that agree with the ground truth. The IG analysis then computes attribution scores for each protein word paired with a privileged fragment. Finally, to obtain complementary pairing rules, protein words with high attribution scores are identified as key contributors to binding activity and are subsequently paired with corresponding privileged fragments. Specifically, for a given privileged fragment, we iteratively select the protein words with the highest positive attribution scores until their cumulative contribution exceeds 50% of the total positive attribution. This automated criterion allows the model to distill the most salient, contributing protein words for each privileged fragment to form the final pairing rules. (Fig. 1a)

To generate protein words for rule prediction, Protein Wordwise[28] was used to partition all protein sequences in the merged dataset (from PDBbind[29], BindingDB[30], BindingNet[31], and ChEMBL[32]) into a set of protein words. ESM-2[33] was applied to compute embeddings for the constituent residues, the average of which served as the embedding for each word (Fig. 1b). The words were subsequently filtered with a preconstructed dictionary to retain high-frequency words occurring across multiple protein sequences. This use of protein words rather than individual amino acids represents the primary difference in feature extraction between PWRules and other common ESM-2-based models.

To obtain small molecule fragments for rule prediction, the MacFrag[34] algorithm was employed to decompose small molecules from the merged dataset into known synthetic drug building blocks (Fig. 1c), which were then filtered to eliminate low-frequency fragments or those with overly simple structural motifs. Physicochemical characterization aimed at assessing the quality of the resulting library of 4,876 drug-like fragments (see Methods for details) revealed that 90.3% fully complied with the "Rule of Three[35]" (Fig. S1a). Additionally, we evaluated the chemical space covered by our fragment library, with coverage rate defined as the proportion of molecules in a given database that contain at least one fragment from our library. Testing on the FDA-approved drug bank, HMDB[36] metabolite database, ChEMBL[32] ligand set, and a ZINC[37] drug-like subset showed coverage rates of 95.2%, 92.9%, 98.3%, and 97.2%, respectively (Fig. S1b). Further analysis of chemical space coverage with tSNE distribution maps confirmed that our fragment library provides comprehensive coverage of drug-like chemical space (Fig. S1c).

After generating protein word–fragment rules with the IG algorithm, we compared these rules with whole protein sequences in both the training and validation sets and computed the accuracy of each rule to assess its reliability and potential applicability to novel proteins. This analysis showed that rules with higher model prediction scores and higher attribution scores were generally more accurate. To provide a metric for rules that reflects both model confidence in the fragment and importance of the protein word for interaction with ligands, we computed rule scores as the geometric mean of the model prediction value and the IG attribution score (Fig. S2). By framing binding prediction as a multi-label privileged fragment prediction task and incorporating an IG-based interpretability module, PWRules thus generates interpretable rules for guiding drug discovery. The final pairing rules library comprises 2,523,287 protein word–fragment rules, encompassing 80,468 unique protein words.

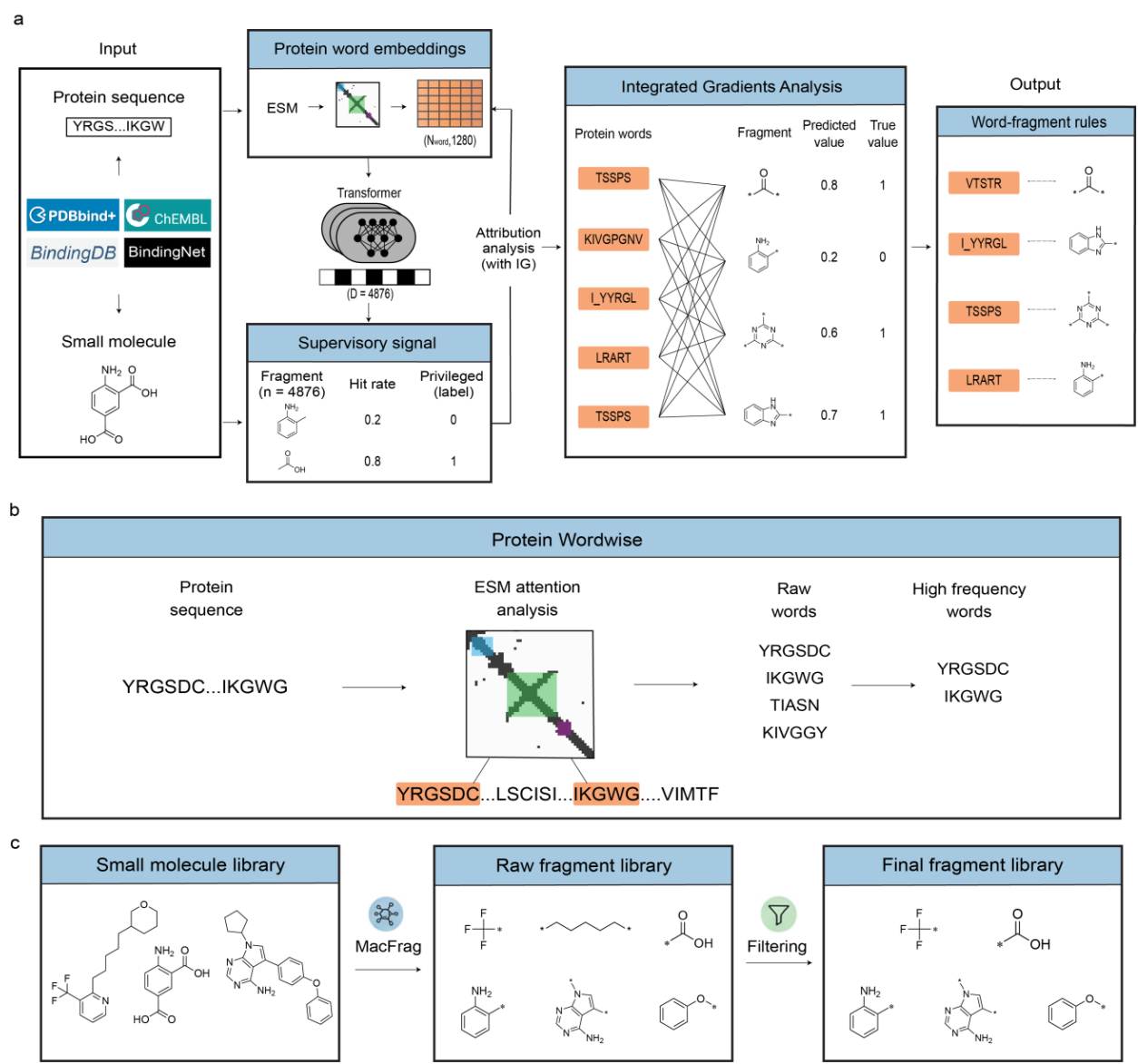

**Figure 1 | The PWRules framework. a**, Overview of the PWRules framework. PWRules takes protein–small-molecule binding data as inputs, then outputs word–fragment rules, each consisting of a protein word paired with a small-molecule fragment. An Integrated Gradients method is applied to compute attribution scores between each protein word and individual privileged fragments. Protein words with high attribution scores are designated as key contributors to ligand interactions and combined with corresponding fragments to form rule pairs. **b,** Workflow for

protein word extraction with Protein Wordwise. **c,** Workflow for extracting small-molecule fragments using the MacFrag algorithm. Finally, a library of drug-like fragments is generated by retaining fragments with a frequency greater than 0.1% and filtering out structurally redundant or undesirable fragments, such as flexible chains with excessive rotatable bonds.

**PWRules identifies privileged small-molecule fragments for diverse drug targets**

To assess the informative value of rules in protein or ligand prediction for drug discovery, we first examined the accuracy of privileged fragment prediction. An input protein sequence is first parsed into protein words using Protein Wordwise and then queried against the rule database to identify matching small-molecule fragments. Each small molecule fragment can potentially match multiple protein words (Fig. 2a), and fragments matching larger numbers of protein words tend to yield higher accuracy predictions (Fig. S3a), we estimated the probability that fragments were correctly designated as privileged by aggregating the scores of all matched rules. Comparison of maximum, averaging, and joint probability approaches for scoring privileged fragments indicated that joint probability formulation achieved the highest accuracy (AUC = 0.775) and was therefore selected as the scoring function for predicting privileged fragments (Fig. S3b).

To evaluate predictive performance in scenarios involving unknown proteins, ligands, or protein–ligand combinations that were absent in model training, we constructed the Novel Protein, Novel Ligand, and Novel Complex datasets, which respectively correspond to absent proteins with ligands provided in the training set, absent ligands with proteins in the training set, and protein–ligand pairs that are both excluded in the training set. We compared the precision of a prediction approach that directly applies the learned rules against two supervised learning models: a Word-embedding-based model (the original supervised model within PWRules) and a Residue-embedding-based model (using raw ESM-2 embeddings as input features). The results revealed that the rule-based approach consistently outperformed both deep learning models across all test sets, achieving precisions of 0.809, 0.866, and 0.772 on the Novel Protein, Novel Ligand, and Novel Complex datasets, respectively. In comparison, the Word-embedding model achieved 0.671, 0.825, and 0.680, while the Residue-embedding model achieved 0.668, 0.802, and 0.689 (Fig. 2b).

To further assess the overall discriminative ability of each method in both positive and negative sample prediction, we calculated the Matthews correlation coefficient (MCC) for samples in each test. The Word-embedding model respectively achieved MCC values of 0.213, 0.518, and 0.189 on the three test sets, the Residue-embedding model achieved values of 0.191, 0.478 and 0.196, compared to values of 0.171, 0.487, and 0.163 achieved by the rule-based method (Fig. 2c). These results suggest that rule-based prediction favors precision, at the cost of reduced coverage across negative sample space, thus supporting the value of integrating rule-based predictions as a complementary module within other models.

To further validate the use of extracted rules in predicting ligands of a real-world drug target, we conducted a representative case study on the non-receptor tyrosine kinase, BTK_HUMAN, which participates in B-cell development, differentiation, and signal transduction, and serves as an important therapeutic target for multiple diseases[38]. For this protein, we examined four known high-affinity ligands, including N-(2-chloro-6-methylphenyl)-2-[[6-[4-(2-hydroxyethyl)piperazin-1-yl]-2-methylpyrimidin-4-yl]amino]-1,3-thiazole-5-carboxamide(Cpd1, K$d$ = 1 nM), 1-[4-[[[6-amino-5-(4-phenoxyphenyl)pyrimidin-4-yl]amino]methyl]piperidin-1-yl]propan-1-one (Cpd2, K$d$ = 2 nM), (2R)-2-[(3R)-3-[4-amino-3-(4-phenoxyphenyl)pyrazolo[3,4-d]pyrimidin-1-yl]piperidine-1-carbonyl]-4,4-dimethylpentanenitrile (Cpd3, IC50 = 4.6 nM), and 7-cyclopentyl-5-(4-phenoxyphenyl)pyrrolo[2,3-d]pyrimidin-4-amine (Cpd4, IC50 = 4.7 nM).

Through our rule-matching strategy, we successfully identified several high-confidence privileged fragments for BTK_HUMAN, including frag_16 (rule score = 0.909), frag_17 (rule score = 0.912), frag_450 (rule score = 0.922), frag_1117 (rule score = 0.963), frag_2045 (rule score = 0.987), and frag_4604 (rule score = 0.987). Notably, all four high-affinity ligands contained multiple privileged fragments: frag_1117 and frag_2045 in Cpd1; frag_16 and frag_17 in Cpd2; frag_17 and frag_4604 in Cpd3; and frag_17 and frag_450 in Cpd4. These results showed that protein word–fragment rules could effectively identify key chemical fragments involved in target binding. As these fragments were recurrently present in well-established high affinity ligands, this case study confirmed that extracted binding rules could guide lead compound discovery and optimization.

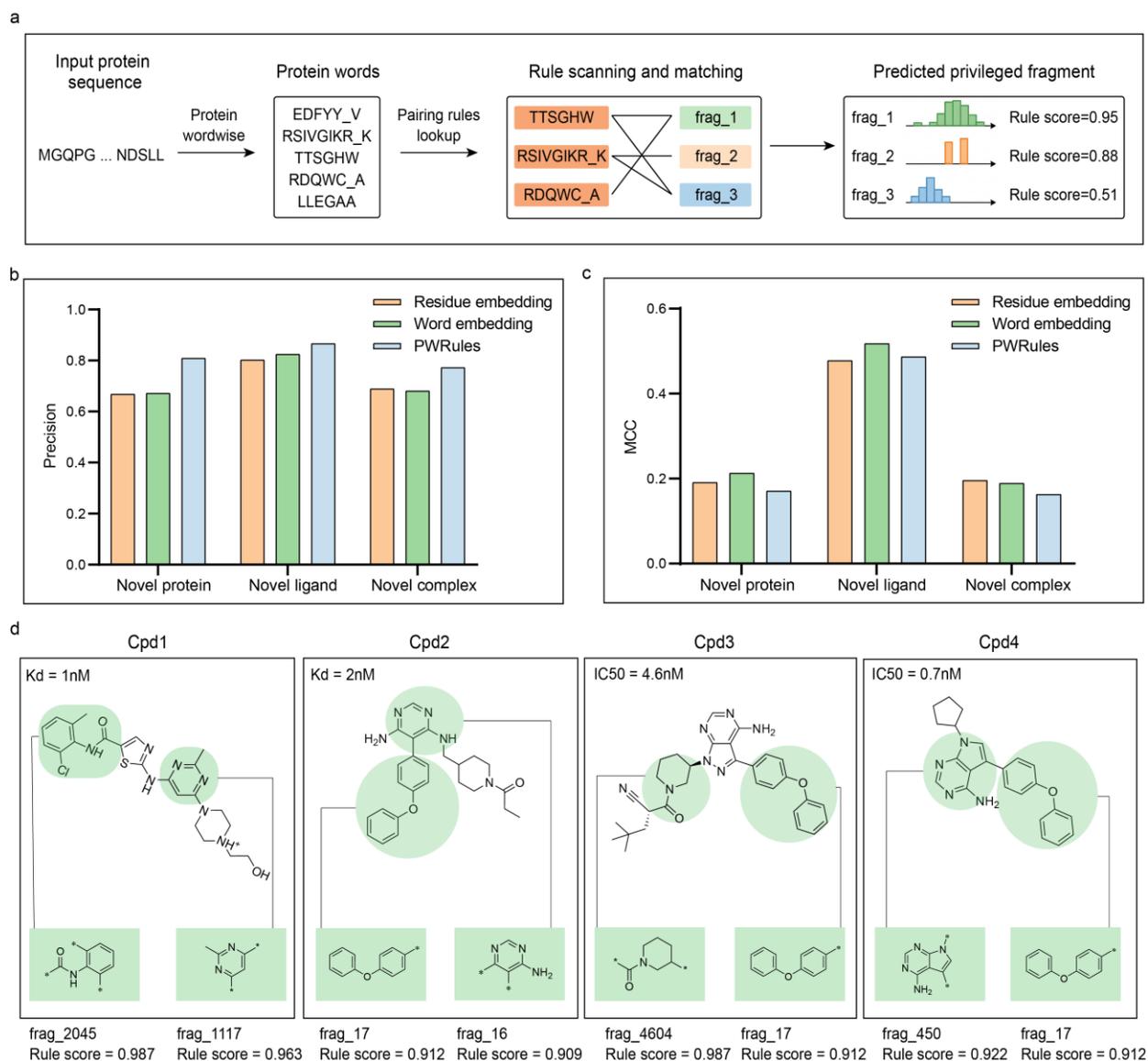

**Figure 2 | PWRules performance in predicting privileged small-molecule fragments for diverse targets. a,** Workflow for privileged fragment prediction with PWRules. Protein words are extracted from the input sequence and sequentially scanned to identify all matching rules in the PWRules database. A prediction score of each fragment's likelihood of being privileged is calculated based on the number of matched rules and their corresponding rule scores. **b-c,** Comparison of precision (b) and MCC (c) among the Residue Embedding baseline model, the Word Embedding baseline model and PWRules across three test datasets (Novel Protein, Novel Ligand, Novel Complex). **d,** Case study of privileged fragments for BTK_HUMAN. Four representative high-affinity small molecules known to bind BTK_HUMAN are shown with high-

scoring privileged fragments predicted by PWRules. Notably, each high-affinity ligand contains at least two of the predicted privileged fragments.

**PWRules encodes interpretable interactions between protein words and small-molecule fragments**

Considering the interpretability of rules obtained by PWRules, we hypothesized that protein word–fragment pairs predicted by our model would be in close spatial proximity within the structure of protein-ligand complexes, consistent with structurally meaningful interactions. To test this hypothesis, we examined three-dimensional structures of protein–ligand complexes from the PDBbind database. After extracting protein words from the protein sequences and small molecule fragments from the ligands of each complex structure, we calculated the distance between the centroids of the protein word and the fragment for every word–fragment pair co-occurring within that complex (Fig. 3a). We found that 47.58% of word–fragment pairs in the training set proteins and 51.37% of pairs in test set proteins were located within 15 Å of each other. To assess whether the close spatial proximity these word-privileged fragment pairs significantly differed from the distribution of random word-fragment pairs, we constructed an equal-sized control set of word–fragment pairs by randomly sampling protein words and molecular fragments. Analysis of word-fragment distance distributions indicated that only 34.49% of random pairs in the training set and 29.95% of the test set were located within 15 Å, both significantly lower than the corresponding proportions of word–privileged fragment pairs predicted by PWRules (Fig. S4). Statistical analysis confirmed this difference: rule pairs were significantly closer than random pairs in both the training (median = 15.4 vs. 18.0 Å) and test sets (median = 14.8 vs. 19.2 Å) (Mann–Whitney U test, $p <$ 0.0001 for each). These results demonstrate that the spatial proximity of rule-based pairs is non-random and structurally meaningful. Although not all predicted pairs were located in close proximity, this substantial proportion of short-distance pairs indicated that many interactions identified through PWRules were indeed structurally plausible and that fragments in these pairs were likely in or otherwise impacted protein regions involved in ligand binding.

To investigate the detailed interactions between protein words and small-molecule fragments, we first examined the PDB structure, 6DH0, as a representative case study, which includes the HIV-1 protease in complex with its high binding affinity inhibitor, Darunavir ([(3aS,4R,6aR)-2,3,3a,4,5,6a-hexahydrofuro[2,3-b]furan-4-yl] N-[(2S,3R)-4-[(4-aminophenyl)sulfonyl-(2-methylpropyl)amino]-3-hydroxy-1-phenylbutan-2-yl]carbamate, K$i$ = 0.026 nM). Structural analysis showed that the backbone of residue G27 forms a hydrogen bond with the nitrogen atom of a peptide bond in the inhibitor (Fig. 3b). Consistent with this observation, our rule database contained a high-scoring pairing rule between the protein word DTGAD (residues 25–29) and the inhibitor fragment frag_2565, which includes this peptide bond (rule score = 0.459). In the complex structure, DTGAD and frag_2565 shared a centroid distance of 5.63 Å, confirming their close spatial proximity and supporting their ability to undergo a direct hydrogen bonding interaction. Given that the DTGAD–frag_2565 rule extracted by PWRules corresponds to a key hydrogen bond in the protein-ligand structure, this case highlights the biological interpretability of our method.

As an additional case study, we examined the PDB structure, 2CEJ, comprising the HIV-1 protease in complex with another high affinity inhibitor, 1AH (methyl N-[(2S)-1-[2-[(2S)-2-benzyl-2-hydroxy-3-[[(1S,2R)-2-hydroxy-2,3-dihydro-1H-inden-1-yl]amino]-3-oxopropyl]-2-[(4-bromophenyl)methyl]hydrazinyl]-3,3-dimethyl-1-oxobutan-2-yl]carbamate, K$i$ = 2.4 nM). Structural analysis revealed an electrostatic interaction between the sidechain of residue D25 and a nitrogen atom in the inhibitor (Fig. 3c). In our rule database, we found a pairing rule (rule score = 0.583) between the protein word, LDTGADDTV (residues 24–32), and a corresponding inhibitor fragment, frag_2279. In the complex structure, LDTGADDTV shared a centroid distance of 7.08 Å with frag_2279, consistent with the close proximity required for this electrostatic interaction, thus demonstrating that pairing rules could capture charge-based interactions. Notably, in this PDB structure, the rule database also identified the aforementioned DTGAD–frag_2565 pairing, although structural analysis revealed a different interaction mode in this context, wherein residue D29 forms a hydrogen bond with the oxygen atom of the peptide bond in frag_2565. These observations suggested that PWRules could identify word–fragment pairs that undergo distinct atomic-level interactions depending on structural context.

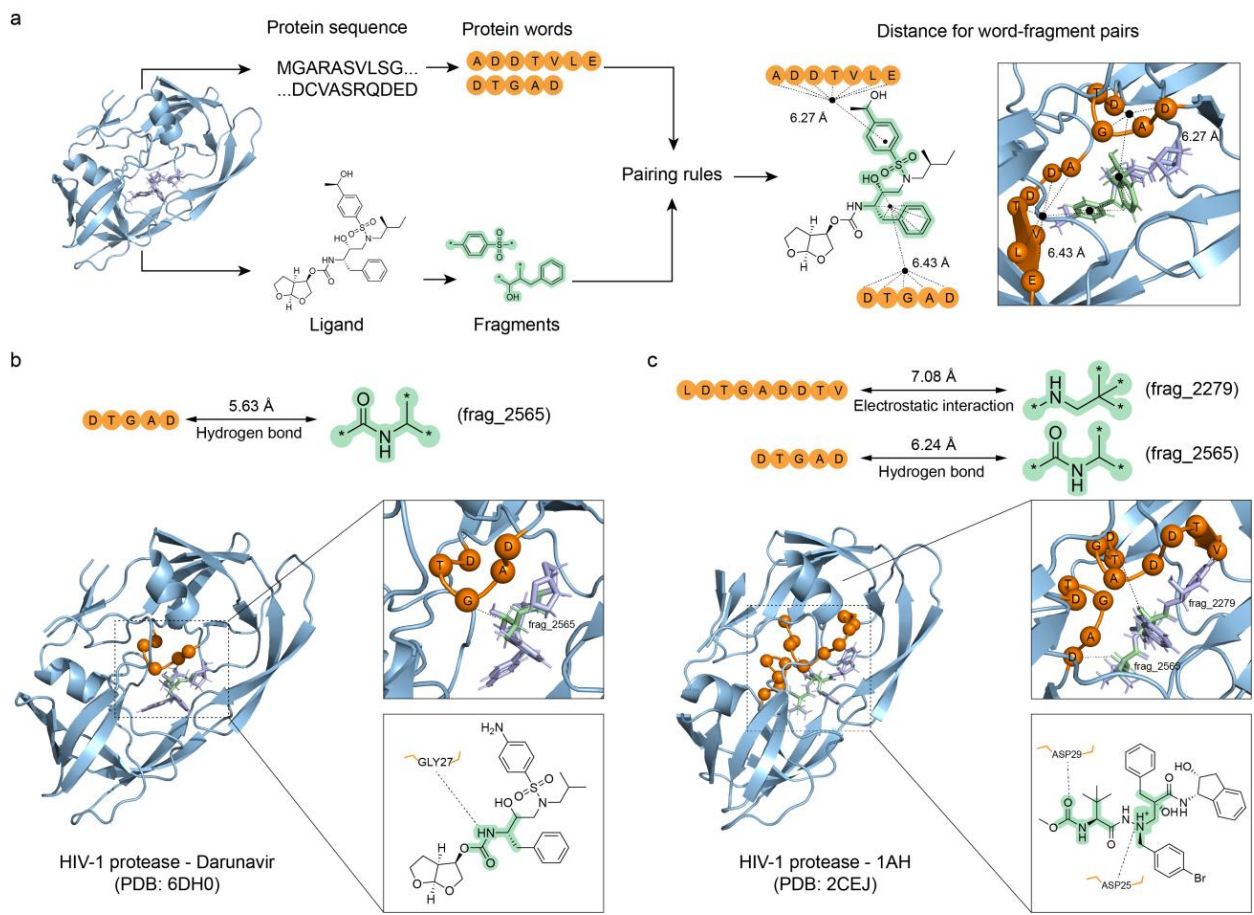

**Figure 3 | Structural validation of protein word–fragment rules using PDB complexes.**

**a,** Method for validating protein word–fragment rules in PDB structures. PDB was scanned for all occurrences of each pairing rule; distances were then calculated between the centroids of each protein word and paired fragment. **b-c,** Two examples illustrating word–fragment rules capturing hydrogen-bonding and electrostatic interactions between protein words and small-molecule fragments. (b) Case study 1: A hydrogen-bonding interaction captured by PWRules (PDB: 6DH0). The high-affinity inhibitor Darunavir binds to HIV-1 protease. PWRules identified a pairing rule between the protein word DTGAD (residues 25–29) and the fragment frag_2565 within the ligand. In the complex structure, their centroid distance is 5.63 Å, consistent with the observed hydrogen bond between the backbone of Gly27 (in DTGAD) and a peptide bond nitrogen in frag_2565. (c) Case study 2: An electrostatic interaction and a hydrogen bond captured by PWRules (PDB: 2CEJ). The inhibitor 1AH binds to HIV-1 protease. PWRules identified a pairing rule between the protein word LDTGADDTV (residues 24–32) and frag_2279 (centroid distance: 7.08 Å), corresponding to a charge interaction between Asp25 and a nitrogen atom in the inhibitor. The rule pairing DTGAD with frag_2565 (same as in d) was also observed, but here it manifests as a distinct hydrogen bond between Asp29 and an oxygen atom in frag_2565.

**PWScore achieves competitive performance in active molecule enrichment**

Building on the ability of PWRules to accurately predict privileged fragments for protein words, we further developed PWScore, a virtual screening method that prioritizes leads based on privileged fragments. PWScore quantifies the degree of matching between candidate small molecules in the screening library and the privileged fragment library predicted by PWRules, with paired candidate small molecules scored and ranked according to their predicted binding propensity to guide target-oriented molecular design (Fig. 4a). The core assumption of PWScore is that small molecules containing a larger number of privileged fragments, especially high confidence fragments, are more likely to undergo binding with protein words in the target. Based on this assumption, we constructed a composite scoring function for small molecules that integrate the number of target-matched privileged fragments within a molecule and rule scores of those corresponding fragments. Specifically, as detailed in the Methods, PWScore evaluates a candidate molecule by identifying privileged fragments that match the target protein. A comprehensive score for each fragment is calculated by multiplying its binding confidence score (derived from the number of associated rules and their scores) by its specificity score. The final PWScore for a molecule is computed by summing the comprehensive scores of these covered fragments, with a maximum atom coverage limit applied to prevent score inflation. The resulting score reflects the overall likelihood of a small molecule binding to a query protein, with higher scores indicating higher probability of activity.

To evaluate the performance of this screening workflow, we compared PWScore with representative physics-based (Glide[39]) and deep-learning–based (PSICHIC[40]) methods on two virtual screening benchmark datasets, including VSDS-vd RandomDecoy (68 protein targets; active compound to random decoy ratio = 1:20), and VSDS-vd MassiveDecoy dataset (8 protein targets; active-to-decoy ratio = 1:300)[41]. Using average enrichment factors (EF) from the top 0.5%, 1.0%, and 5.0% of the ranked lists as evaluation metrics, we found that PWScore achieved EF values of 13.4, 11.3, and 5.5 on the RandomDecoy dataset, respectively. In comparison, Glide achieved EF values of 12.1, 10.7, and 6.0, while PSICHIC achieved EF values of 12.6, 10.8, and 5.8 (Fig. 4b). Similarly, PWScore obtained EF values of 40.2, 24.2, and 7.2 on the MassiveDecoy dataset, whereas Glide yielded EF values of 38.4, 24.0, and 8.0, PSICHIC yielded EF values of 34.2, 23.4, and 7.4 (Fig. 4c). These results indicate that PWScore achieves comparable enrichment performance to both physics-based and deep-learning-based methods while additionally providing interpretability through explicit privileged fragment mapping.

Notably, PWScore captures rule-based interaction features distinct from the energy calculations employed by Glide and the complex pattern recognition features employed by PSICHIC. Therefore, we hypothesized that PWScore could serve as a complementary module to enhance the predictive accuracy of these established screening methods. To explore this, the raw scores from PWScore and the baseline models (Glide or PSICHIC) were first individually transformed using Z-score normalization, and the average of these standardized scores was used to rank the compounds. This combined strategy resulted in superior performance. On the RandomDecoy dataset, the combination of Glide and PWScore achieved EF values of 15.5, 13.5, and 7.2, while the combination of PSICHIC and PWScore reached EF values of 15.5, 13.4, and 6.8. Consistent improvements were also observed on the MassiveDecoy dataset (Fig. 4b-c). These findings demonstrate that by integrating the interpretable, rule-based insights of PWScore with the rigorous scoring of established models, we can achieve a synergistic effect that greatly improves active molecule enrichment.

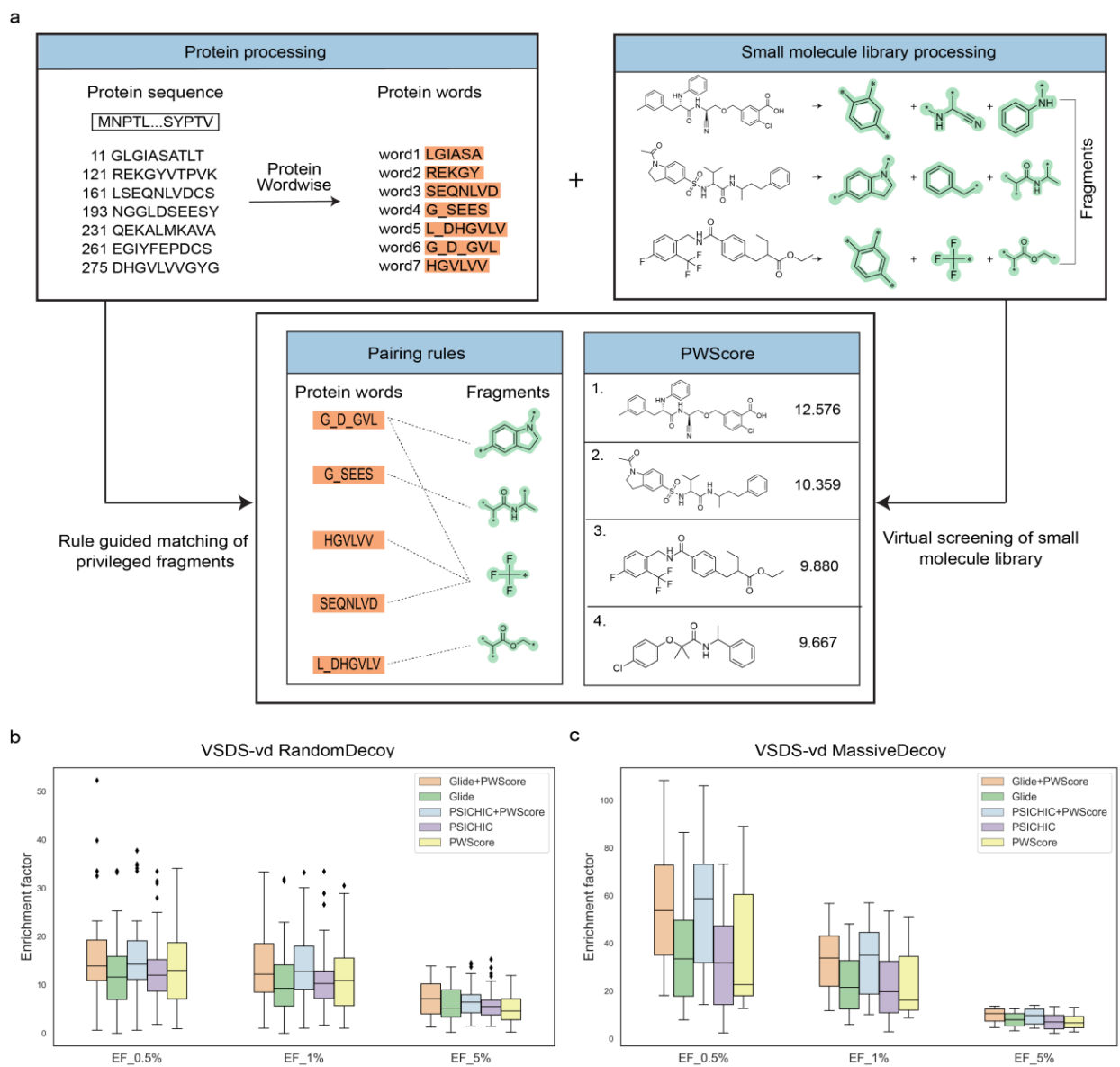

**Figure 4 | Evaluation of PWScore in virtual screening tasks.**

**a**, Workflow of virtual screening using PWScore. Privileged fragments are first predicted for the target protein using PWRules, after which PWScore is used to score and rank compounds in the screening library based on the predicted privileged fragments. **b-c**, Enrichment factor performance of different methods on the VSDS-vd RandomDecoy dataset (**b**) and VSDS-vd MassiveDecoy dataset (**c**) at EF0.5%, EF1%, and EF5%.

**Proof-of-principle demonstration of PWScore generalizability and application in real-world virtual screening tasks**

For further proof-of-principle evaluation of the practical application and generalizability of PWRules and PWScore in drug discovery, we conducted a case study assessing our framework's performance in virtual screening for ligands of the primary SARS-CoV-2 main protease, M$^{pro}$ (also known as 3CL$^{pro}$), a key therapeutic target in anti-COVID-19 drug development required for viral replication. To this end, we compiled an extensive independent test set of M$^{pro}$ binding data encompassing 15,843 small molecules from the literature and public databases, 12.7% of which are known active molecules (Fig. 5b). Importantly, M$^{pro}$ was not included in training data for either PWRules or PWScore, ensuring stringent evaluation of generalizability to previously unseen targets (Fig. 5a). In this independent evaluation, the combination of Glide and PWScore achieved an EF0.5% score of 6.77 (Fig. 5c), compared to 6.48 for Glide alone. Notably, in this combined approach, putative active molecules accounted for 86.3% of the top 0.5% ranked compounds, substantially exceeding enrichment through random selection. These results indicated that integrating PWScore with established methods could effectively discriminate active from inactive compounds, even for protein targets not encountered during training, thus demonstrating its generalizability and application in real-world virtual screening tasks.

Using the combined Glide and PWScore approach, we accurately identified Nirmatrelvir (IC50 = 2.67 nM, score = 3.38, rank = 76), the active component of the FDA approved drug, Paxlovid, as well as its covalent inhibitor, Boceprevir (IC50 = 4.13 μM, score = 2.96, rank = 123). To understand how PWScore contributed to the improved ranking of these actives, we next investigated whether word-fragment rules reflected physical binding principles of M$^{pro}$ inhibitors using the structure of Nirmatrelvir in complex with M$^{pro}$ (PDB: 7TLL) as an example. This analysis revealed that PWRules identified an interaction between the M$^{pro}$ protein word, "ELPTGV" (residues 166–171), and the small molecule fragment, frag_2027 (Fig. 5d), with a hydrogen bond observed between the backbone of E166 and the peptide bond nitrogen in frag_2027. This finding suggested that PWScore reflected binding characteristics in the vicinity of this word. Additionally, we examined the structure of M$^{pro}$ (PDB: 7NBR) in complex with Boceprevir. PWRules identified an interaction between the protein word "GTTTLN" (residues 23–28) and fragment frag_4730, which covers the amide nitrogen of Boceprevir through hydrogen bonding, explaining the binding characteristics of Boceprevir. These results showed that word–fragment rules for these compounds encode physical principles of binding, enabling the model to prioritize active compounds.

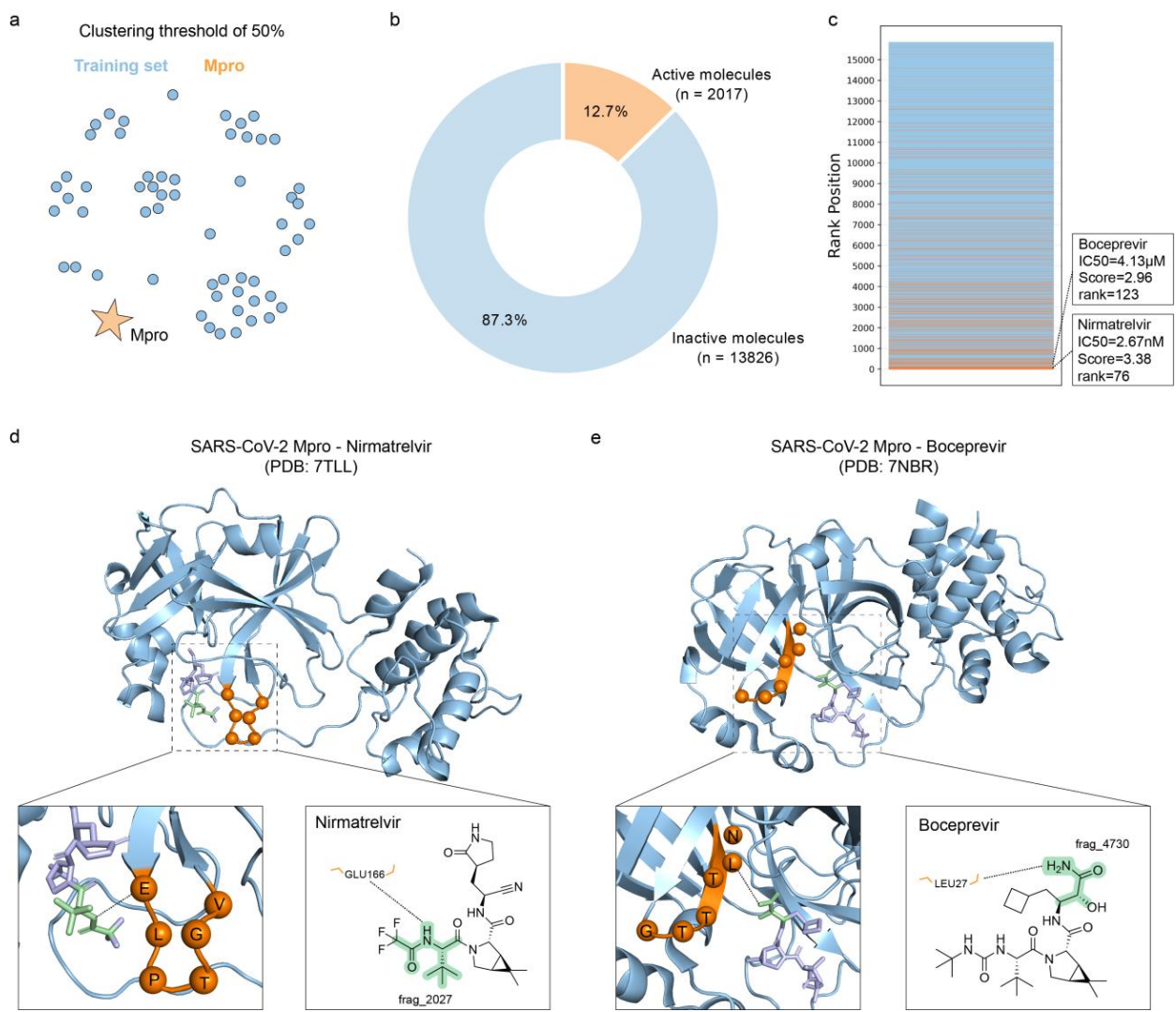

**Figure 5 | Inhibitor prediction for SARS-CoV-2 Main Protease using PWScore to augment Glide with word-fragment rules. a,** Target analysis showing M^pro is excluded from the training dataset. **b**, Proportion of active molecules in the M^pro test set. **c**, Ranking of compounds in the M^pro test set by the combination of Glide and PWScore. **d-e**, Case studies of M^pro inhibitors, Nirmatrelvir (d) and Boceprevir (e), to visualize matched protein word–fragment rules in complex crystal structures for analysis of possible binding mechanisms.

**DISCUSSION**

In this study, we introduce the PWRules framework for predicting complementary pairing rules between protein words and small-molecule fragments. PWRules leverages binding affinity data to predict privileged small molecule fragments for a given protein and employs an interpretability module to translate these predictions into word-fragment rules. Building on these rules, we developed the PWScore function for prioritizing active compounds, which demonstrated competitive performance in active molecule enrichment compared to other scoring functions while uniquely offering interpretability. We found that PWScore is broadly generalizable to proteins outside the training set, such as the SARS-CoV-2 main protease. Although trained without explicit structural guidance, our analyses of validated protein-ligand complexes showed that PWRules learns protein word–fragment rules that are enriched near ligand-binding pockets (Fig. 3). Additionally, PWRules and PWScore are highly adaptable, and can be integrated with other deep learning-based methods to improve predictive accuracy in drug discovery tasks. By extracting and applying these complementary pairing rules, PWRules offers an interpretable framework for enhancing current computational drug discovery methods, bridging deep learning models with foundational principles of protein-ligand interactions.

Historically, heuristic rules have been widely used in drug design efforts[4], wherein pairing rules are typically applied to the protein active site, while non–active site residues that might also contribute to ligand binding are often ignored[42-44]. In addition, medicinal chemists frequently focus on residue-substrate interactions, but overlook cooperative interactions among residues, especially among residues outside the active site. Our present framework enables prediction of higher-order sequence patterns involved in ligand recognition by applying protein words to capture local sequence context with functional relevance, consequently extending the rule set beyond active-site regions and single–residue-level knowledge. Through protein words, PWRules can model cooperative effects among residues that are often missed by amino acid–level embeddings. The learned rules also provide an interpretable knowledge base for drug discovery, which can be applied at multiple stages of drug design, including (i) prioritization of fragment libraries for fragment-based drug discovery, (ii) de novo molecule design targeting specific protein words, and (iii) predicting active molecules for sequence-only targets lacking structural information, which are particularly relevant for targets harboring intrinsically disordered regions.

Additionally, we noted that rules involving words located outside a binding pocket might reflect long-range or pre-binding interactions required for ligand recognition[45]. Given that some rules are not yet fully understood, especially those involving protein words outside of active sites, we proposed a cascade model for understanding how pairing rules predicted by PWRules and PWScore could exhibit high prediction accuracy (Fig. S5)[46]. Briefly, growing evidence suggests that high-affinity ligands do not necessarily localize to protein binding pockets through purely random three-dimensional diffusion, but instead proceed through initial, transient associations with the protein surface mediated by long-range electrostatic interactions and shallow hydrophobic patches. This process, commonly referred to as electrostatic steering or encounter complex formation, selectively enriches ligands near functional regions of the protein, effectively increasing local ligand concentrations near the binding site. Two-dimensional surface diffusion and/or local conformational rearrangements subsequently guide ligands into deeply buried binding pockets[47]. Our model can thus capture protein words outside the active site that play a role in ligand enrichment, guidance, and stabilization during the binding process, supporting the utility of pairing rules that do not involve the binding pocket.

Beyond the word-fragment rules described in the current study, future work could extend the PWRules framework to other classes of biomolecular interactions, including protein–protein interaction (PPI) rules[48,49] and protein–nucleic acid (NA) pairing rules[50,51]. This expanded capacity would require construction of PPI or protein–NA interaction datasets to substitute the protein–small molecule binding datasets employed in our present work. Although the current overall rule structure would remain similar for PPIs, with protein words defined on both interacting partners, to accommodate protein–NA interactions, our framework would require development of a dedicated motif dictionary defining the sequences or structural motifs of DNA and RNA. Despite these challenges, this adaptability to other basic research and drug design questions highlights the generalizability of our framework to diverse protein–ligand binding systems beyond small molecule recognition.

## METHODS

**Dataset Construction and Data Preprocessing**

*Data Sources*

Protein-small molecule affinity data were collected from four databases: PDBbind[29], BindingDB[30], BindingNet[31], and ChEMBL[32]. In these databases, proteins with sequence lengths exceeding 1024 amino acids were excluded because such sequences are computationally demanding to process. A binding event was defined as a protein–ligand pair with experimentally determined Ki, Kd, IC50 or EC50 < 10 μM. For affinity data from the ChEMBL database, entries with assay type annotated as "Binding" or "Functional" were retained while protein complexes entries were removed. During data integration, duplicate records were processed based on the following priority order: PDBbind > BindingDB > BindingNet > ChEMBL Binding > ChEMBL Functional, while the affinity type priority was Kd > Ki > IC50 > EC50. For repeated measurements of the same affinity type, the median value was selected as the final affinity value. All affinity data were binarized using a 10 μM threshold (values <10 μM were considered active). The final dataset contained 9,661 proteins, 1,635,634 small molecules, and 3,438,736 affinity data points.

*Fragment Library Generation*

Small-molecule fragmentation was performed using the MacFrag[34] algorithm. Fragments appearing in >50% of binding-positive ligands were retained as 'privileged', while fragments that occurred in < 0.1% of small molecules were removed. The resulting library was further refined by filtering out structurally redundant or undesirable substructures, such as flexible chains with excessive rotatable bonds and complex linear peptides. The final library contained 4,876 drug-like fragments.

To evaluate the chemical space coverage of our molecular fragment library, the proportion of molecules was computed in four benchmark databases: FDA-approved Drug Library, HMDB[36] metabolite database, ChEMBL ligand set and the ZINC[37] drug-like subset. Given the massive scale of the ZINC database (ZINC20 contains over 200 million drug-like molecules), 2 million molecules were randomly sampled from the ZINC20 drug-like subset to strike a balance between computational efficiency and representativeness for coverage calculations.

*Protein Word Extraction*

Protein sequences were partitioned into biologically meaningful semantic units, termed "Protein Words," through the application of Protein Wordwise[28], an unsupervised segmentation toolkit based on the attention mechanisms of Protein Language Models (PLMs). Unlike conventional segmentation strategies that rely on rigid k-mers or sliding windows, our strategy parsed sequences according to intrinsic residue dependencies captured by the pre-trained ESM-2 model[33].

In this procedure, attention matrices were first extracted from the transformer layers of the ESM-2 model for each input sequence. These matrices, which reflect pairwise interactions between amino acid residues, were utilized to construct residue-interaction graphs where nodes represented residues and edges represented attention weights. Subsequently, the Louvain community detection algorithm was applied to these graphs to identify clusters of highly correlated residues, which were then designated as protein words.

To ensure the functional relevance of the segmented units, a filtering criterion was imposed: only segments containing 5 to 20 amino acid residues were retained. This specific length range was selected to align with the typical dimensions of functional motifs and structural domains commonly observed in nature. It should be noted that protein words defined in this manner are not restricted to contiguous substrings; instead, discontinuous residues that are spatially or functionally linked in the primary sequence were also grouped into individual semantic units, thereby enabling the capture of long-range dependencies.

For the vectorization of these units, the hidden state representations from the ESM-2 model were employed. Each protein word was represented by a fixed-length vector, which was computed by averaging the embeddings of all constituent residues within that word. Through this approach, the biochemical context and evolutionary information of each functional segment were effectively aggregated.

*Dataset Annotation and Splitting*

A 'privileged fragment' for each protein is defined as the molecular fragments that appeared in >50% of its binding ligands with a binding affinity < 10 μM. This yielded a binary matrix of 9,661 proteins × 4,876 fragments, where each entry indicated whether a fragment is privileged for a given protein. This matrix served as the supervisory signal for model training. To evaluate the generalization performance of the model, the dataset was partitioned into training, validation, and test sets using an 8:1:1 ratio. The validation and test sets were constructed under three complementary settings: novel proteins, novel ligands and novel complexes. All splits were generated prior to training, ensuring no data leakage during the training process.

**The PWRules Framework Architecture**

*Base Model Architecture*

The PWRules model adopted a Transformer encoder architecture for multi-label prediction of privileged fragments to capture the contextual interactions between protein words and molecular fragments. The model took protein word embeddings as input, outputting a multi-dimensional vector (length 4,876), where each dimension corresponded to the predicted probability of a fragment being privileged for the input protein.

*Input Representation*

Each protein word was represented as a fixed-length embedding vector (1 × 1280), where 1280 is consistent with the embedding dimension produced by ESM-2. A learnable classification token (CLS) was prepended to the sequence of protein words to capture global contextual information. The final hidden state of this token was used as an aggregated protein-level representation for downstream fragment prediction. The fragment space was defined as a fixed library of molecular fragments extracted from ligand SMILES strings.

*Training Protocol*

PWRules was trained using Adam optimization with an initial learning rate of $1\times10^{-3}$, a weight decay of $1\times10^{-5}$, and a batch size of 256. A cosine annealing learning rate scheduler (T_max = 20) was employed. Model parameters were initialized with fixed random seeds for reproducibility.

Each training run spanned up to 600 epochs. Early stopping was applied if validation performance did not improve for 60 consecutive epochs. Model checkpoints with the highest average Mathews correlation coefficient (MCC) across independent validation sets were selected as final weights. The model was trained using binary cross-entropy with logits loss (BCEWithLogitsLoss). To address missing labels (NaN), the loss values were masked, and normalization was performed only over the observed entries.

**Interpretability Module and Rule Extraction**

*Attribution Analysis*

The Integrated Gradients (IG)[52] attribution method was used to interpret the predictions of the model, enabling the identification of protein words that contributed to predictions. As a gradient-based explainability approach, Integrated Gradients attributed a model's prediction to its input features by integrating gradients along a path from a baseline input to the actual input. This step was implemented using the Captum[53] library. The attributions in PWRules were computed using a forward function by reconstructing the transformer layers, attention masks, layer normalization and the final multilayer perceptron used for prediction. Integrated Gradients were computed independently for each fragment prediction by specifying the corresponding output neuron as the attribution target. The resulting attribution tensor was condensed to a single contribution score per protein word by summing across the embedding dimension. Thus, a single contribution score per protein word was obtained which was normalized using the L2 norm. Attribution analysis was performed exclusively on positive fragment interactions via the CPU.

*Rule Scoring Definition*

The Rule Score is defined as the geometric mean of the model prediction score and the IG attribution score. This combined score reflects the confidence and strength of the rule, with higher scores indicating greater confidence.

$$Rule\ Score = \sqrt{Prediction\ Score \times Attribution\ Score}$$

*Rule Filtering*

The accuracy of each PWScore was evaluated in both the training set and validation set. Rules with an accuracy below 0.5 were excluded, thereby ensuring that the remaining rules were robust and generalizable.

**Construction of PWScore for Virtual Screening**

*Prediction of Privileged Fragments*

The PWScore function was developed based on the protein word-fragment rules extracted from the PWRules. To predict privileged fragments for a target protein, its amino acid sequence was first segmented into protein words using Protein Wordwise. These words were then scanned against the precomputed rule database to identify potential privileged fragments that might bind to the protein. The binding confidence for each fragment was calculated using a joint probability formula, which integrated the number of pairing rules associated with the fragment and their respective rule scores. The underlying principle was that a higher number of rules pointing to the same fragment, coupled with elevated rule scores, increased the reliability of the fragment's binding propensity. The confidence score is defined as follows:

$$S_{conf}(f) = 1 - \prod_{i=1}^{n}(1 - R_i)$$

Here, $S_{conf}(f)$ denotes the confidence score of a specific privileged fragment $f$. The variable $n$ represents the total number of rules associated with this fragment, while $R_i$ corresponds to the rule score of the i-th rule.

Additionally, the specificity of a privileged fragment was evaluated based on its frequency of occurrence across ligands in the dataset. Fragments with higher occurrence frequencies are generally simpler and more generic but exhibit lower target specificity. To quantify this, a specificity score is introduced, computed as:

$$S_{spec}(f) = 1 - \frac{L_f - L_{min}}{L_{max} - L_{min}}$$

Here, $L_f$ denotes the logarithmic frequency of fragment $f$. $L_{min}$ and $L_{max}$ denote the minimum and maximum logarithmic frequencies observed across all fragments in the dataset, respectively.

The foundational assumption of PWScore is that candidate molecules containing a larger number of privileged fragments with high confidence and specificity scores are more likely to bind to the target protein. Thus, the comprehensive score for each privileged fragment is defined as the product of its confidence score and specificity score:

$$S_{comp}(f) = S_{conf}(f) \times S_{spec}(f)$$

*Scoring Function Formulation*

During virtual screening, PWScore evaluates a candidate molecule by first identifying all privileged fragments within the molecule that match the target protein. These fragments are then covered in descending order of their comprehensive scores, and the scores of the covered fragments are summed to compute the total PWScore for the molecule. A higher PWScore indicates a greater probability of the molecule being active:

$$PWScore(M) = \sum_{f \in F_{matched}} S_{comp}(f)$$

In this expression, $PWScore(M)$ is the aggregate predicted activity score for a candidate molecule $M$. The summation is performed over $F_{matched}$, which denotes the set of unique privileged fragments identified within the molecule. Each fragment contributes its individual comprehensive score, $S_{comp}(f)$, to the total.

To prevent score inflation caused by overlapping low-score fragments, a maximum atom coverage limit is implemented. Through parameter tuning, we determined that setting the maximum coverage count to 10 per atom optimizes PWScore's performance in molecular ranking tasks.

**Structural Validation of Pairing Rules**

**Geometric Analysis in PDB**

To assess the structural plausibility of predicted protein word–fragment pairs, a geometric analysis was performed using experimentally resolved protein–ligand complexes from the PDBbind database. For each complex structure, spatial relationships between co-occurring protein words and molecular fragments were evaluated by computing centroid-based distances. The centroid of a protein word was calculated as the geometric center of the Cα atoms in its representative residues while the centroid of a molecular fragment was defined as the geometric center of the three-dimensional coordinates of its constituent heavy atoms. Euclidean distances between centroids were computed for all co-occurring protein word–fragment pairs to quantify spatial proximity and assess the structural consistency of predicted pairing rules.

Specific molecular interactions (hydrogen bonds, electrostatic contacts) between protein words and privileged fragments were identified using Schrödinger's Protein Interaction Report tool. Hydrogen bonds were detected based on the software's built-in geometric criteria for donor–acceptor pairs, while electrostatic interactions were assessed via spatial analysis of charged groups.

**Virtual Screening Benchmarks**

*Baseline Methods Setup*

For comparative evaluation, we employed two baseline virtual screening methods: Glide[39] and PSICHIC[40].

Glide was configured using a docking grid centered on the centroid of the reference ligand in each holo protein structure. The grid box dimensions were set to 16 Å plus 0.8 times the diameter of the co-crystallized ligand, ensuring comprehensive coverage of the binding site. Docking scores were computed using Glide's standard precision mode.

PSICHIC was implemented using the official codebase, retrained on our custom dataset to align with the study's objective. All other parameters were retained as defaults to maintain consistency with the original methodology.

*Evaluation Metrics*

Model performance was assessed using the enrichment factor (EF), which quantifies the concentration of active molecules within a ranked subset relative to random selection. The EF at a given threshold (e.g., top 1%) is defined as:

$$EF_{x\%} = \frac{Hits_{x\%}/N_{x\%}}{Hits_{total}/N_{total}}$$

This metric evaluates the early enrichment capability of virtual screening methods.

*Decoy Datasets*

Benchmarking utilized two decoy datasets: VSDS-vd[41] RandomDecoy (active-to-decoy ratio = 1:20) and VSDS-vd[41] MassiveDecoy (ratio = 1:300). To ensure consistency with the PWRules framework, an activity threshold of 10 μM was applied uniformly for defining active molecules and corresponding decoys in both datasets.

## Data Availability

Our code is available at https://github.com/TianBoxue-lab/PWRules.

## Author Contributions

J.C. developed PWRules and PWScore. J.Z. performed protein word extraction using Protein Wordwise. T.T. prepared the figures. Z.S. implemented the PSICHIC baseline. X.Z. implemented the Glide baseline. B.T. designed the project, wrote and revised the manuscript.

## Acknowledgments

This work was supported by Beijing Frontier Research Center for Biological Structure (No. 041500002), Tsinghua University Initiative Scientific Research Program (No.20231080030), and the Tsinghua-Peking University Center for Life Sciences (No.20111770319).

## Competing interests

The authors declare no competing interests.

## Author Information

Correspondence and requests for materials should be addressed to B.T. (boxuetian@mail.tsinghua.edu.cn).

# Supplementary Information

## An Interpretable Framework Applying Protein Words to Predict Protein–Small Molecule Complementary Pairing Rules

Jingke Chen[1*], Jingrui Zhong[1*], Tazneen Hossain Tani[1*], Zidong Su[1*], Xiaochun Zhang[1*], Boxue Tian[1]

[1] MOE Key Laboratory of Bioinformatics, State Key Laboratory of Molecular Oncology, Beijing Frontier Research Center for Biological Structure, School of Pharmaceutical Sciences, Tsinghua University, Beijing, 100084, China

* These authors contribute equally to this work.

[#] To whom correspondence should be addressed:

Boxue Tian: boxuetian@mail.tsinghua.edu.cn

## Table of Contents

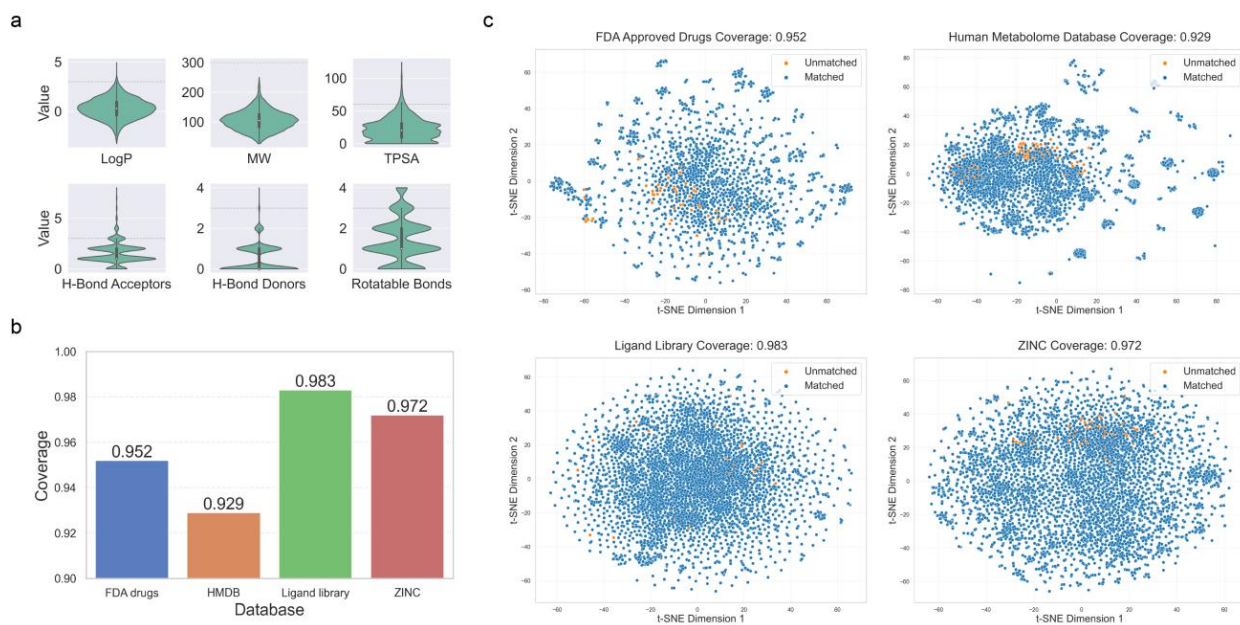

**Fig. S1 | Fragment library physicochemical properties and chemical space coverage analysis. a,** Distributions of physicochemical properties: LogP, molecular weight (MW), polar surface area (TPSA), hydrogen bond acceptors, hydrogen bond donors, and rotatable bonds of the refined fragment library. **b,** Coverage of fragment library across different compound databases- FDA-approved drugs (95.2%), HMDB metabolites (92.9%), ligand library (98.3%) and ZINC drug-like molecules (97.2%). **c,** Coverage of fragment library shown through t-SNE distribution of molecular fragments across databases. Blue indicates molecules covered and orange indicates molecules not covered by the fragment library.

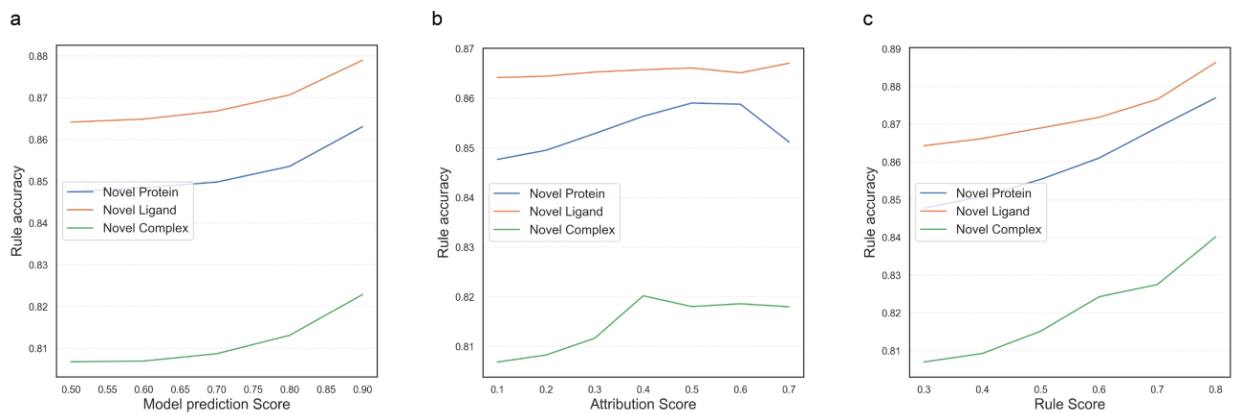

**Fig. S2 | Rule accuracy as a function of key metrics. a,** Relationship between model prediction score and rule accuracy. **b,** Relationship between attribution score and rule accuracy. **c,** Relationship between rule score (the geometric mean of the model prediction and attribution scores) and rule accuracy. The three curves in each panel correspond to the three test scenarios: Novel Protein, Novel Ligand, and Novel Complex. Overall, higher model prediction scores, attribution scores, and rule scores are associated with greater rule accuracy, indicating that the rule score effectively integrates model confidence and feature importance to identify more reliable protein–fragment association rules.

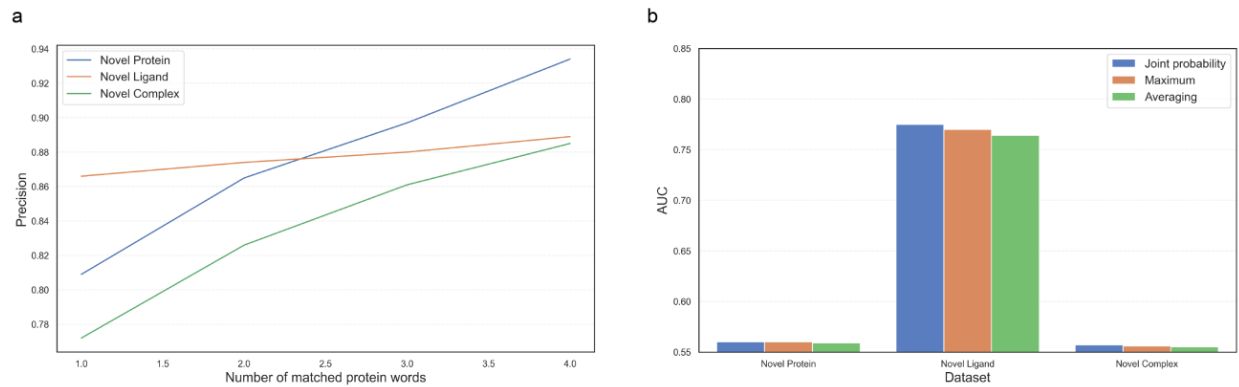

**Fig. S3 | Performance of predicting privileged fragments based on pairing rules. a,** Relationship between precision and the number of matched protein words. Lines represent the three test sets: Novel Protein, Novel Ligand, and Novel Complex. Overall, precision systematically increases as the number of protein words matched by a fragment grows, indicating that multi-word matches provide stronger predictive signals and higher reliability. Although the exact values vary slightly across test sets, all show a consistent positive correlation. **b,** Comparison of AUC values for different privileged fragment scoring functions across the three test sets. The bar plot shows AUC values calculated using three scoring strategies— "Joint probability," "Maximum," and "Averaging"—for the Novel Protein, Novel Ligand, and Novel Complex test sets. In all cases, the "Joint probability" method achieves the highest AUC, demonstrating its superiority as a scoring function.

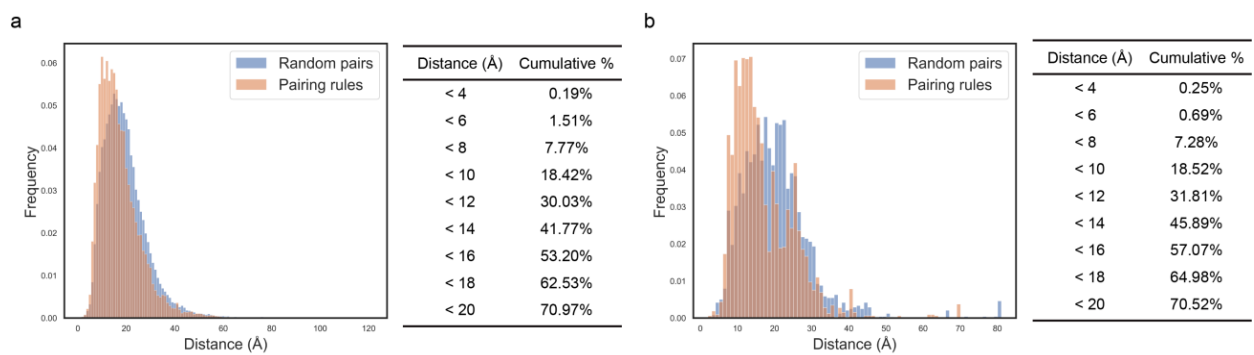

**Fig. S4 | Structural validation of protein word–fragment pairing rules in training and test sets. a,** Distance distributions between protein word–fragment pairs identified by PWRules (orange) versus randomly sampled pairs (blue) in PDB structures from the training set (left panel), with cumulative percentages of word-fragment pairs within specified distance thresholds shown in the right panel. **b,** Corresponding distance distributions and cumulative percentages for the test set.

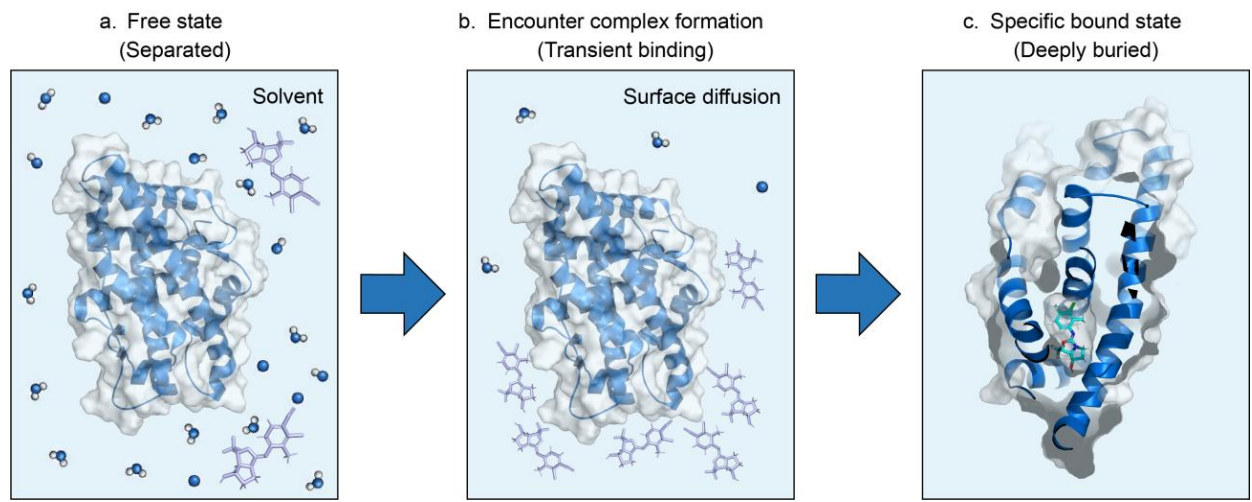

**Fig. S5 | Cascade model for protein–ligand binding mechanism. a,** Free state where protein and ligand are separated in bulk solution. **b,** Encounter complex formation via electrostatic steering and surface diffusion, where ligands transiently associate with the protein surface before reaching the binding site. **c,** Specific bound state with the ligand deeply buried in the protein binding pocket.